\documentclass{article}

\usepackage{arxiv}

\usepackage[utf8]{inputenc} 
\usepackage[T1]{fontenc}    
\usepackage{hyperref}       
\usepackage{url}            
\usepackage{booktabs}       
\usepackage{amsfonts}       
\usepackage{nicefrac}       
\usepackage{microtype}      
\usepackage{lipsum}
\usepackage{graphicx}  
\usepackage[ruled,vlined]{algorithm2e}
\usepackage{amsmath}
\usepackage{amssymb}  
\usepackage{float}

\title{Consolidation via Policy Information Regularization\\ in Deep RL for Multi-Agent Games}

\author{
  Tyler Malloy, Chris R. Sims \\
  Department of Cognitive Science\\
  Rensselaer Polytechnic Institute\\
  Troy, NY 12180 \\
  \texttt{\{mallot,simc3\}@rpi.edu} \\
   \And
 Tim Klinger, Miao Liu, Matthew Riemer, Gerald Tesauro \\
  IBM Research AI \\
  Yorktown Heights, NY 10598 \\
  \texttt{\{tklinger, mdriemer, gtesauro\}@us.ibm.com} \\
  \texttt{\{miao.liu1\}@ibm.com}  \\
}

\begin{document}
\maketitle

\begin{abstract}
This paper introduces an information-theoretic constraint on learned policy complexity in the Multi-Agent Deep Deterministic Policy Gradient (MADDPG) reinforcement learning algorithm. Previous research with a related approach in continuous control experiments suggests that this method favors learning policies that are more robust to changing environment dynamics \cite{malloy2020deep}. The multi-agent game setting naturally requires this type of robustness, as other agents' policies change throughout learning, introducing a nonstationary environment. For this reason, recent methods in continual learning are compared to our approach, termed Capacity-Limited MADDPG. Results from experimentation in multi-agent cooperative and competitive tasks demonstrate that the capacity-limited approach is a good candidate for improving learning performance in these environments. 
\end{abstract}


\section{Introduction}
In the current paper we explore a \textit{capacity limited}\footnote{This objective is also referred to as the mutual-information regularized learning objective in \cite{leibfried2020mutual}.} approach that explicitly constrains the amount of information used to represent an RL policy \cite{leibfried2020mutual} by effectively constraining the difference between the agent's policy and marginal action distribution. Previous research using this approach demonstrated improved generalization in continuous control environments \cite{malloy2020deep}. However, the precise differences that this regularization method induces on learning remains unclear. This paper seeks to extend the understanding of the impact that this regularization technique has on learning by applying onto learning in multi-agent environments with mixed co-operative and competitive games. 

Multi-agent games introduce a fundamental nonstationarity to the environment, due to changes in other agents' behaviour over the course of learning. This nonstationarity is similar to the domain of continual learning, in which environment dynamics may change over time, introducing a challenge of learning new skills and maintaining old ones, referred to as the \textit{stability-plasticity} trade-off. One previous approach to dealing with this challenge is known as policy consolidation \cite{kaplanis2019policy}; this is a method of improving continual learning by constraining the difference as measured by KL-divergence between the current policy and previous policies.

The policy consolidation and capacity-limited methods share the approach of constraining the difference between the current policy and some baseline. However, neither has previously been applied to the domain of multi-agent games. This work seeks to understand to connection between consolidation and generalization of learned behaviour in environments with changing dynamics. Specifically, we investigate the impact of applying a policy information constraint onto the domain of mixed cooperative-competitive games with $n >= 2$ agents. This is done by altering the traditional reinforcement learning objective to include a regularization based on the amount of information being used to represent and agent's policy. 

\section{Background}
\subsection{Multi-Agent Reinforcement Learning}
The Multi-Agent Reinforcement Learning (MARL) setting comes in co-operative, competitive, and mixed cooperative/competitive game variations. In the experimentation portion of this paper, the novel capacity-limited version of MADDPG will be compared with the standard method in each of these types of environments. MARL methods can be distinguished as either training independent learners, which do not model other agent's policies, and joint action learners, which do attempt to model these policies \cite{claus1998dynamics}. The MADDPG method, and by extension the capacity-limited version, does attempt to represent the policy of other agents in the learning environment.

An important aspect of the MARL setting is the requirement of agents to respond to changing behaviour of other agents within the environment. This is particularly true of games with some aspect of competitiveness as failure to appropriately adapt to the changing policies of opponents can result in opponents learning to take advantage of a too stable policy. This failure of plasticity can also impact the speed of learning in strictly cooperative games, as failing to appropriately account for the changing behaviour of compatriots can slow learning. 

Another important goal of the MARL setting is preventing the forgetting of previously learned behaviour or knowledge of the learning environment. In the competitive setting, forgetting optimal behaviour in response to certain behaviour of enemy agents can result in those agents learning to take advantage of forgotten behaviour. This aspect of MARL is the closest relation to the continual learning setting, where dynamics of the environment change overtime, with the ultimate goal of learning to be improved performance across the future experience within this changing environment. 

\subsection{Continual Reinforcement Learning}
The challenge of continual learning is represented by an agent that must continually improve their behaviour throughout their lifetime while acting in an environment with changing dynamics \cite{parisi2019continual}. These changing environment dynamics introduce competing objectives for the agent, optimizing on current environment dynamics at each time step and optimizing performance in previous states of the environment. Failure to properly balance this trade-off leads to \textit{catastrophic forgetting} in which agents learn only to optimize performance in relation to recently experienced environment dynamics, and would perform poorly if reintroduced to previously experienced environment dynamics. 

Continual RL can be related to the MARL setting by noting that other agents introduce changing environment dynamics through the updating of their learned behaviour. Under this conceptualization, catastrophic forgetting represents a failure to act appropriately in relation to previously observed behaviour of other agents. This failure can be in relation to the behaviour of either opponents or compatriots, as both types of agents potentially impact learning environment dynamics.
\section{Capacity-Limited Method}
The capacity-limited learning objective imposes a limitation on the amount of information that is used to represent an agent's policy, as defined by the mutual information of the policy $\mathcal{I}(\pi(a|s)$ according to the objective:
\begin{equation}
\begin{split}
\text{max}_{\pi_{0:T}} \mathbb{E}_{(s_t,a_t) \sim \rho_{\pi}}\bigg[ \sum_{t=0}^T r(s_t,a_t) \bigg]  \text{ s.t } \mathbb{E}_{(s_t,a_t) \sim \rho_{\pi}}[\mathcal{I}(\pi(a|s)) \leq \mathcal{C}] \text{  } \forall t
\end{split}
\end{equation}
where $\mathcal{I}(\pi(a|s))$ is the mutual information of the policy function when taken to be the information channel mapping states onto actions. The precise method for approximating this quantity will be discussed in the section on capacity-limited MADDPG. This allows us to define $\mathcal{C}$, the desired maximum channel capacity, and optimize performance in the environment in relation to this capacity. In practice, this is done by applying a weighted regularizer term to the reinforcement learning objective based on the mutual infomration of the agent's policy. This optimization can be used to define a learning objective that better reflects the reality of information constraints on physical agents, and as we will see, has theoretic connections to the method of policy consolidation when applied to continual learning environments. 

As noted previously, this policy mutual information regularization based approach has been shown to improve performance in some continual control tasks in simulated robotics environments \cite{leibfried2020mutual}. However for the purposed of this paper, the interesting aspect of this approach is further research applying this regularization method which has shown improved generalization and robustness within these same robotics simulation environments when altering physics simulation parameters \cite{malloy2020deep}. The learning that takes place when altering these environment parameters is closely related to continual learning, meaning it may have an impact on learning that takes place in the MARL setting due to the connection between these.  

\subsection{Capacity-Limited Learning Objective}
The connection of the policy information capacity with consolidation and generalization will be motivated more thoroughly in later sections, but the intuitive justification is that policies that are simpler in an information-theoretic sense discourage the over-fitting of earlier experience in environments with changing dynamics. In practice, the way we impose a limit on the amount of information that the agent uses to represent its policy is done by applying a penalty to the reward based on this value. This allows us to define a learning objective that regularizes the observed reward:
\begin{equation*} 
J(\pi) = \sum_{t=0}^T \mathbb{E}{(s_t,a_t) \sim p_{\pi}}[r(s_t,a_t) - \beta \mathcal{I}(\pi(\cdot|s_t)] 
\end{equation*}
\begin{equation*} 
J(\pi) = \sum_{t=0}^T \mathbb{E}[r - \beta \mathcal{I}(\pi(A|S)] 
\end{equation*}
The key difference with the standard RL objective is the added penalty to the reward observed based on the amount of information that would be required to represent the policy. Policies with higher mutual information values have a greater complexity, in an information-theoretic sense, and this weighted value is used to discourage policies that would require a high information capacity channel. Thus, this learning objective will directly encourage the development of policies that are simple (use low information to represent) but have high utility. Additionally, if there are multiple policies that achieve the same performance, this objective will naturally favor the simplest among them. Higher values of $\beta$ skew the learning objective to prefer policies with less required information.

The limitation that is imposed on the information capacity of the agent's policy is introduced by this learning objective. Because this learning objective is used to update the agent's policy throughout training, the information capacity of the learned policy will be dependent on the value of $\beta$. In the extreme, very high values of $\beta$ will train an agent to prefer a policy that requires as little information to represent as possible over any improvement in the reward. Because of the nature of information capacities, this policy could be either uniform in all states of the environment and perform actions randomly everywhere, or deterministic in all states or perform the same action everywhere. Conversely, setting the value of $\beta$ to zero results in the traditional learning objective of maximizing the reward with no limitation on the amount of information utilized by the policy. 

\section{Related Work}
\subsection{Relation to MADDPG}
The MADDPG method \cite{lowe2017multi} serves as the baseline for performance in multi-agent mixed cooperative-competitive games that will be built upon using the capacity-limited approach. As mentioned, one of the motivating factors for applying capacity-limits onto the multi-agent domain is the potential benefit for nonstationary problems that it can afford. However, the MADDPG method does have aspects that seek to address this issue. 

Firstly, MADDPG employs a policy ensemble method that randomly selects a policy from a list of potential policies to use to define the action of an agent at any time-step. This is done to prevent the over-fitting of learned behaviour onto the actions of another agent in the environment, which will likely change over time. Additionally, this issue of nonstationarity introduced by other agent's learning altering their behaviour is partially mitigated in the MADDPG model by utilizing a mix of centralized and decentralized learning. The decentralized nature of this learning ensures that at execution time actions can be performed without information from other agents, but improves learning speed by sharing a centralized action-value function $Q_i^{\mu}$ which is updated as:
\begin{equation}
\begin{split} \label{eq:CentralizedActionValueFunction}
    \mathcal{L}(\theta_i) = \mathbb{E}_{x,a,r,x'}[(Q_i^{\mu}(x, a_1, ... , a_N) - y)^2], 
    y = r_i + \gamma Q_i^{\mu'}(x', a_1', ..., a'_N) |_{a'_j=\mu)j(a)j),}
\end{split}
\end{equation}
\subsection{Relation to Policy Consolidation RL}
The policy consolidation method attempts to improve performance in nonstationary learning environments by keeping a history of recent policies and incorporating them into the learning objective of the agent. This is done to prevent the forgetting of useful behaviour that was beneficial to the agent in previous time steps of the environment. This results in the following learning objective:
\begin{equation*}
\begin{split}
\mathcal{L}^*(\pi) = \mathcal{L}(\pi_1) + \mathbb{E}_{s_t \sim \rho_1} \Big[ \sum_{k=1}^{N-1} g_{k, k+1}  \mathcal{D}_{KL}(\pi_k || \pi_{k+1}) \Big]
\end{split}
\end{equation*}
This can be connected to the capacity-limited method by noting that it is based on a similar goal of minimizing the loss of the agent's policy, which an additional constraint or regularization that alters this learning objective: 
\begin{equation}
\begin{split}
\label{constrained_optimization_mi}
\textbf{Goal: } \text{Minimize } \mathbb{E}[L(\pi(a|s))]  \text{subject to } \mathcal{I}(\pi(a|s)) \leq \mathcal{C}
\end{split}
\end{equation}
This capacity is achieved in the capacity-limited learning objective by setting the coefficient $\beta$ to weight the mutual information regularization $\mathcal{I}(\pi(a|s))$ such that it applies the policy information capacity $\mathcal{C}$. One interpretation of this regularization term is the KL-Divergence between their policy and the product of the state and action marginals: 
\begin{equation}
\mathcal{I}(\pi(a|s)) = \mathcal{D}_{KL}(\pi(a|s) || p_{\pi(a|s)}(a) \otimes p_{\pi(a|s)}(s)) 
\end{equation}
In this way, the capacity-limited learning objective can be seen as similar to that of the policy consolidation method. Instead of using multiple previous polices as a baseline to limit deviation from, the capacity-limited method approximates the marginal action distribution and penalizes behaviour that varies too much from this approximation. A key aspect of both models is that the regularization term is a weighted penalty on the reward that the agent observes, meaning that there is control over how much deviation from the baseline is allowed. As this weight term goes to zero, both models allow for unlimited deviation from the baseline, and as it increases the deviation is limited. 

The difference between the two methods is that instead of using previous policies $\pi_{k_{\text{old}}}(a|s)$ as is done in the policy consolidation method, the capacity-limited approach uses an approximation of the marginal action distribution $\pi(a)$. Additional differences between the two methods exist in that the capacity-limited method only uses a single marginal, as opposed to a series of previous policies as is done in the policy consolidation method. Together these differences separate the two methods, but their similarities do provide justification for investigating the impact of the capacity-limited method in learning environments that have changing dynamics.
\section{Capacity-Limited MADDPG}
The introduction of the capacity-limited learning objective is done by simply modifying the MADDPG centralized action-value function in Equation \ref{eq:CentralizedActionValueFunction} as follows: 
\begin{equation} \label{eq:CLActionValueFunction}
\begin{split}
    y = (r_i - \beta \mathcal{I}(\pi(a|s)) +   \gamma Q_i^{\mu'}(x', a_1', ..., a'_N) |_{a'_j=\mu)j(a)j),}
\end{split}
\end{equation}
The $\beta$ parameter balances the impact that the reward has on the learning objective with the penalty on the amount of information that is being used to represent the agent's policy. Because this parameter is dependent on the scale of the reward, it needs to be fit to each individual learning environment. In the section on multi-agent environments, results from experimentation are tested using a range of beta parameters, and results are shown with different parameters demonstrating the impact on performance that changing this parameter has. Apart from this alteration of the learning objective, the main difference in performance comes from the method of approximating the mutual information of the agent's policy as is detailed in the following section. 

Before detailing the method of approximating the mutual information of the agent's policy, it is important to understand how to conceptualize this quantity in relation to an agent that acts deterministically. The issue introduced by deterministic policies for calculating the mutual information is that this quantity will always be 0, as the action the agent performs is completely determined by the state they are in. Although the MADDPG algorithm uses a deterministic policy during each episode of training, this policy is randomly drawn from a ensemble of K different policies at the beginning of each episode \cite{lowe2017multi}. 

The result of this is that, taken independently of the random drawing of policies that takes place at the beginning of each episode, there is a degree of randomness in relation to the state that the agent is in and the action they perform. Additionally, as we will see in the next section on approximating policy information, the method of approximation results in a comparison of the entropy of average behaviour across many episodes of the environment, and the current batch of states and actions that is being used to train the agent at each iteration of policy update.

\subsection{Policy Information Approximation}
The goal of the policy information approximation method is to calculate the mutual information of the agents policy as closely as possible. This mutual information term can be calculated using the definition of mutual information in terms of its constituent entropies
\begin{equation} \label{eq:MI}
\mathcal{I}(\pi(a|s)) = \mathcal{H}(\pi(a)) - \mathcal{H}(\pi(a|s)) 
\end{equation}

As mentioned, the method of approximating the mutual information of the policy is unique to the CL-MADDPG method, as the underlying method it is based on is different from the two previous examples in \cite{malloy2020deep} and \cite{leibfried2020mutual}, which are based on the Soft-Actor Critic (SAC) method \cite{haarnoja2018soft}. The reason that MADDPG requires a unique approximation is that the actions taken by the agent are deterministic, whereas actions in the SAC method are defined by a mean and variance.

This definition of actions allowed the capacity-limited version \cite{malloy2020deep} of SAC to directly calculate the entropy of the agent's policy at any specific state, and approximate the marginal action distribution using a diagonal multi-variate Gaussian. This allows us to define the marginal action distribution similarly as $\mathcal{N}(\mu_{\rho}, \sigma_{\rho}^2)$ and calculate the mean and variance as:
\begin{equation*}
\begin{split}
\mu_{\rho} & = \sum_s p(s) \mu_s  \\
\sigma_{\rho}^2 & = \sum_s p(s) \sigma_s^2 + \sum_s p(s) \big( \mu_s \big)^2 - \Big( \sum_s p(s) \mu_s \Big)^2
 \end{split}
\end{equation*}
Where $\mu_s$ is a vector representing the mean of the policy at state s and $\sigma_s^2$ is the variance. As the SAC policy is stochastic, at each step $n$ through the environment the policy outputs a mean $\mu_n$ and variance $\sigma^2_n$ defining the action that will be performed in that state. This gives us the online updating method for the estimate of the marginal action distribution variance $\hat{\sigma}_n^2$ and mean $\hat{\mu}_n$ with the learning rate $\alpha$.
\begin{equation*}
\begin{split}
\hat{\mu}_n       = \alpha \mu_n + (1-\alpha) \hat{\mu}_{n-1} \\
\hat{\sigma}_n^2  = \alpha \sigma_n^2 + (1-\alpha) \hat{\sigma}_{n-1}^2 
                 & + \big( \alpha \mu_n^2 + (1-\alpha) (\hat{\mu}_{n-1})^2 \big) 
                 - \big( \alpha \mu_n + (1-\alpha) \hat{\mu}_{n-1}  \big)^2  
 \end{split}
\end{equation*}
However, this method cannot be used in the MADDPG method because of the deterministic nature of the actions selected by the agents. Instead, the values $\mu_s$ and $\sigma_s^2$ are estimated at each time step based on the mean and variance of the previous 100 actions the agent has performed. After calculating these values, the estimate of the marginal action distribution as a diagonal multi-variate Gaussian is done in the same way as earlier giving the approximations at time n as $\mu_n$ and $\sigma_{\mu}^2$. After approximating these values, the marginal action probability of an action $a$ is calculated in relation to a diagonal multi-variate Gaussian based on these approximations as follows:
\begin{equation*}
\begin{split}
\mathcal{H}(\pi(a)) = - \sum_{a \in D} \mathcal{N}(\hat{\mu}_n, \sigma_{\mu}^2)(a) \log \Big( \mathcal{N}(\hat{\mu}_n, \sigma_{\mu}^2)(a) \Big)
\end{split}
\end{equation*}
Where $D$ is the current batch of environment actions that is being used to train the model. Another issue introduced by the MADDPG model using deterministic actions is that the entropy of the agent's policy cannot be done in the same way as the earlier method. Instead this term is estimated based on calculating the mean $\mu_{D}$ and variance $\sigma^2_{D}$ of the batch and representing the agent's policy entropy as follows:
\begin{equation*}
\begin{split}
\mathcal{H}(\pi(a|s)) = - \sum_{a \in D} \mathcal{N}(\mu_{D}, \sigma^2_{D})(a) \log \Big( \mathcal{N}(\mu_{D}, \sigma^2_{D})(a) \Big)
\end{split}
\end{equation*}
Ultimately, this means that the approximation of the mutual information is based on the difference in entropy between the current batch of actions that is being used to train the model, and the marginal approximation of the likelihood of those actions being performed independent of the state of the environment. 
\subsection{Capacity-Limited MADDPG Algorithm}
The capacity-limited MADDPG algorithm is derived based on the MADDPG algorithm present in \cite{lowe2017multi}, altered to include the steps that approximate the policy mutual information as well as the updated centralized action-value function in Eq \ref{eq:CLActionValueFunction}. The major differences are the steps that approximate the marginal action distribution which is done by updating a window of recent actions taken with a running average of the means and variance of the action window. At each step a new action is added to the window and a mean and variance of the window is used to updated this running average with a small learning rate $\alpha = 0.001$. This gives the running approximation of the marginal action distribution in relation to the diagonal multi-variate gaussian defined by this running average of action mean and variances as $\pi_{\mu}(a_t)$. 

During each training step, the current minibatch of actions is used to define a diagonal multi-variate gaussian, and the probability of each action in the minibatch is calculated in relation to this distribution $\pi_{\kappa}(a_t|s_t)$. Finally, the mutual information of the agents policy is approximated with the entropies of these two probabilities as: $\mathcal{I}(a|s) = \mathcal{H}(\pi_{\mu}(a_t)) - \mathcal{H}(\pi_{\kappa}(a_t|s_t))$. The only remaining step is to add a weighted regularizer that penalizes the magnitude of this mutual information approximation, as is done in Eq \ref{eq:CLActionValueFunction}.

Taken all together, this results in the Capacity-Limited MADDPG algorithm as follows:
\begin{algorithm}
    \For{episode = 1 to M}{
        Initialize a random process $\mathcal{N}$ for action exploration \\
        Receive initial state $x$  \\
        \For{t = 1 to max-episode-length}{
            for each agent i, select action $a_i = \mu_{\theta_i} + \mathcal{N}_t$ w.r.t the current policy and exploration \\
            Execute actions $a = (a_1, ... , a_N)$ and observe reward r and new state $x'$ \\
            Store (x,a,r,x') in replay buffer $\mathcal{D}$ \\
            Add the action a to the action window A, remove oldest action if full \\
            $\pi_{\mu}(a_t) \sim \mathcal{N}(\hat{\mu}_A, \hat{\sigma}_A^2)$ \\
            $x \leftarrow x'$ \\
            \For{agent i = 1 to N}{
                Sample a random minibatch of S samples $(x^j, a^j,r^j,x'^j)$ from D \\
                $\pi_{\kappa}(a_t|s_t) \sim \mathcal{N}(\hat{\mu}_{\mathcal{D}}, \hat{\sigma}_{\mathcal{D}}^2)$ \\
                $\mathcal{I}(a|s) = \mathcal{H}(\pi_{\mu}(a_t)) - \mathcal{H}(\pi_{\kappa}(a_t|s_t))$ \\
                set $y^j = r_i^j - \beta \mathcal{I}(\pi(a|s)) + \gamma Q_i^{\mu'}(x'^j, a'_1,...,a'_N) |_{a_k=\mu_k(o_k^j)}$ \\
                Update critic by maximizing the loss $\mathcal{L}(\theta_i) = \dfrac{1}{S}\sum_j \big(y^j - Q_i^{\mu}(x^j, a^j_1,...,a^j_N)\big)^2$ \\
                Update actor using the sampled policy gradient:
                $$ \nabla_{\theta_i} J \approx \dfrac{1}{S} \sum_j \nabla_{\theta_i} \mu_i(o_i^j) \nabla_{a_i}Q_i^{\mu}(x^j, a_1^j,...,a_i,...,a_N^j) |_{a_i=\mu_i(o_i^j)} $$
            }
            Update target network parameters for each agent i
            $$ \theta'_i \leftarrow \tau \theta_i + (1 - \tau) \theta_i'$$
        }
    }
\caption{Capacity-Limited MADDPG\label{CL-MADDPG}}
\end{algorithm}
\newpage
\section{Mutli-Agent Environments}
Four of the multi-agent environments from the original MADDPG paper \cite{lowe2017multi} are used to compare learning performance of the capacity-limited version against the original. These environments consist of a continuous state and action space with a discrete time representation. Players in these environments act within a 2D grid and move throughout the environment seeking landmarks that are associated with reward. Environments were chosen to form a base of the different types of co-operation or competitiveness, strict co-operation, strict competition, and mixed co-operative/competitive environments. 

As mentioned, the reason for applying the policy mutual information regularizer that defines the capacity-limited approach is in furthering the understanding of how it impacts learning in non-stationary environments. These different MARL environments have different forms of non-stationarity, meaning that this regularization strategy may me more or less applicable onto these environments depending on the type of non-stationarity that is present in that environment. Although the sample of MARL environments used here cannot give a complete picture of how the capacity-limited approach can impact all multi-agent or non-stationary environments, the results presented in the following sections do provide a clearer understanding of its impact.     
\begin{figure*}[ht] 
\begin{centering}
  \includegraphics[width=\textwidth]{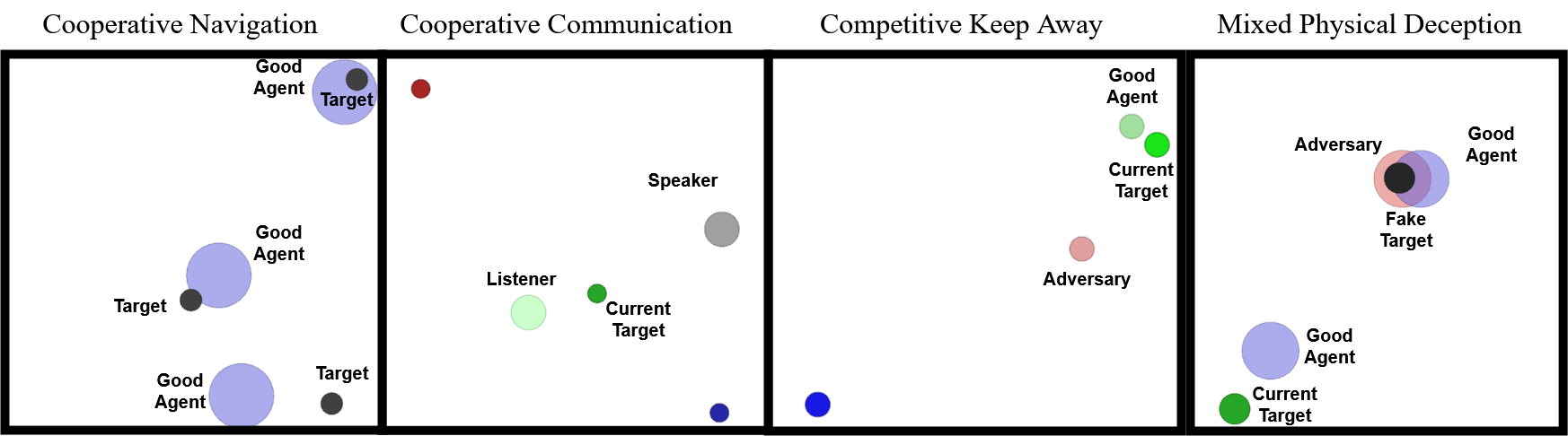}
  \caption{Multi-Agent environments used as a test bed for the capacity-limited version of MADDPG. LEFT: Cooperative Navigation task, 3 good agents (light blue) spread to each of the 3 targets (dark grey). Only covering 1 or 2 targets will not maximize the reward. LEFT-CENTER: Cooperative Communication task, the speaker (Grey) communicates with the listener (light green) which of the three locations (green, red, or blue) is the target for this episode. The listener must move to the target location indicated by the speaker and learn which location corresponds to the information being communicated by the speaker. RIGHT-CENTER: Competitive Keep Away environment in which the adversary (light red) must push the good agent (light green) away from the target location. The adversary does not know at the beginning of an episode where the target location for the good agent is (green or blue) and must infer it from their behaviour, if the good agent reaches the target location before the adversary it will be able to remain there. RIGHT: Mixed cooperative and competitive task where 2 good agents (light blue) move towards the target location (green) and prevent the adversary (light red) from moving there. This can be done by tricking the adversary to move towards the dummy target (black) as the adversary cannot see which of the locations is the target.}
 \end{centering}
\end{figure*}

\subsection{Cooperative Environments}
The first cooperative environment used to compare the traditional and capacity-limited versions of MADDPG is the Cooperative Communication task in which one stationary agent chooses an action that acts as a signal to the other agent which of the 3 potential locations that agent should move towards to maximize their reward. The mobile agent can see when they are near one of the three possible locations, but do not know the color, and must infer which location they should head towards based on the communication from the other agent. 

\begin{figure}[ht] 
    \begin{centering}
    \includegraphics[width=0.75\textwidth]{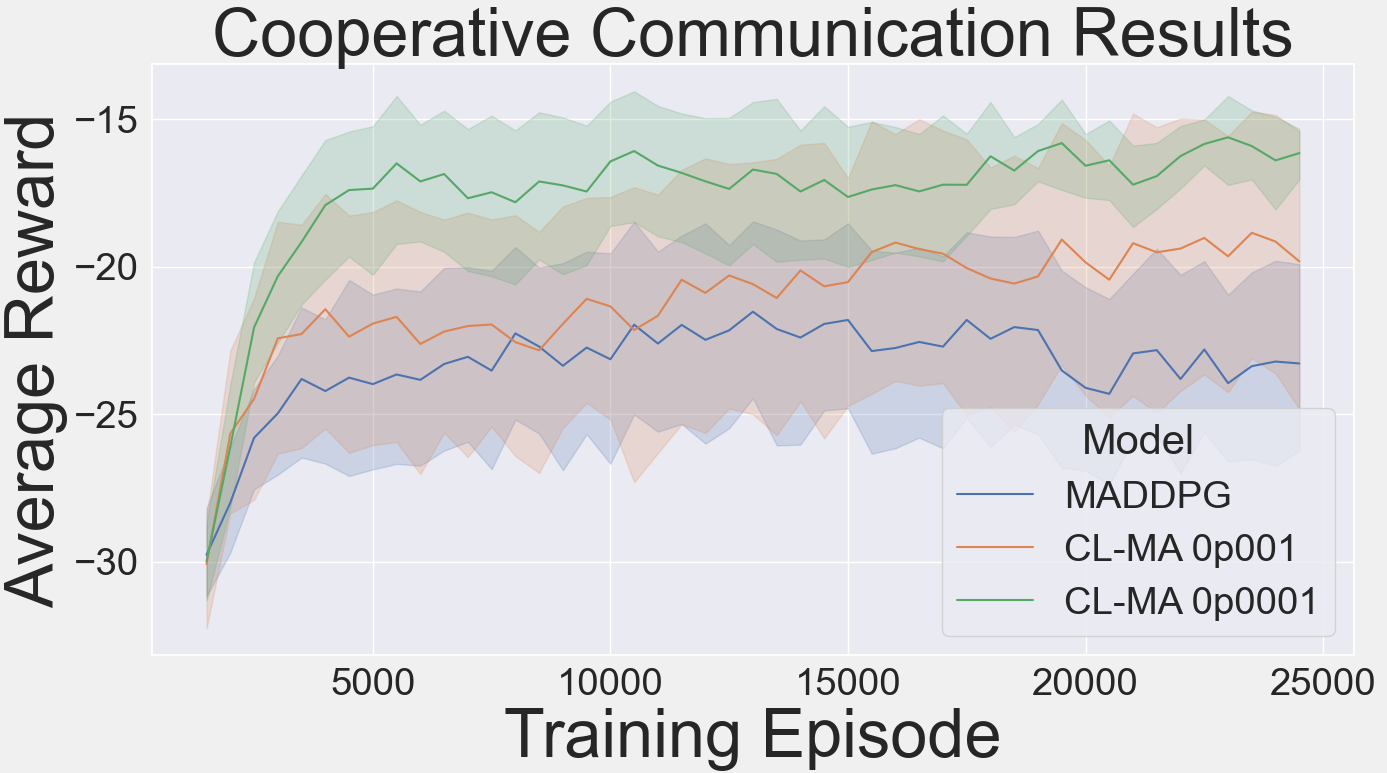}
    \caption{MADDPG and CL-MADDPG (labelled as CL-MA) training results in the cooperative communication environment. Green line shows a CL-MA agent with a $\beta$ coefficient of 1e-3. Orange line shows the same scenario with a coefficient of 1e-2. Blue line shows the traditional MADDPG agent. Averages are shown over 5 seeds, with a rolling average window of 5 episodes used to smooth the curve. Error bars represent 99\% confidence interval.}
    \label{fig:SpeakerListener}
    \end{centering}
\end{figure}

Results from experimentation in the Cooperative Communication task shown in Figure \ref{fig:SpeakerListener} demonstrate that the capacity-limited method with a $\beta$ coefficient of 1e-2 achieves better performance at the end of training compared to the traditional MADDPG method. The larger difference in performance is observed in the agent with a $\beta$ coefficient of 1e-3, which allows for more consistently high reward. 

While the performance variance for the 1e-2 $\beta$ and traditional DDPG agent are fairly large, the 1e-3 $\beta$ has more consistent performance (lower variance across agent seeds). This consistency of learned behaviour matches with the justification for the capacity-limited approach, which attempts to encourage the similarity of new behaviour to the marginal action distribution when it is associated with an increase in reward. 

An additional cooperative environment was used to compare performance between the capacity-limited and standard approach in a learning environment with more than 2 agents. In these types of environments, the nonstationarity introduced for each agent by 2 different updating policies can lead to difficulties in the stability of learning. However, as these results show, the capacity-limited agent with a $\beta$ parameter of 1e-3 is able to learn quickly and with a low degree of sparseness in observed performance across different training seeds. 
\begin{figure}[ht] 
    \begin{centering}
    \includegraphics[width=0.75\textwidth]{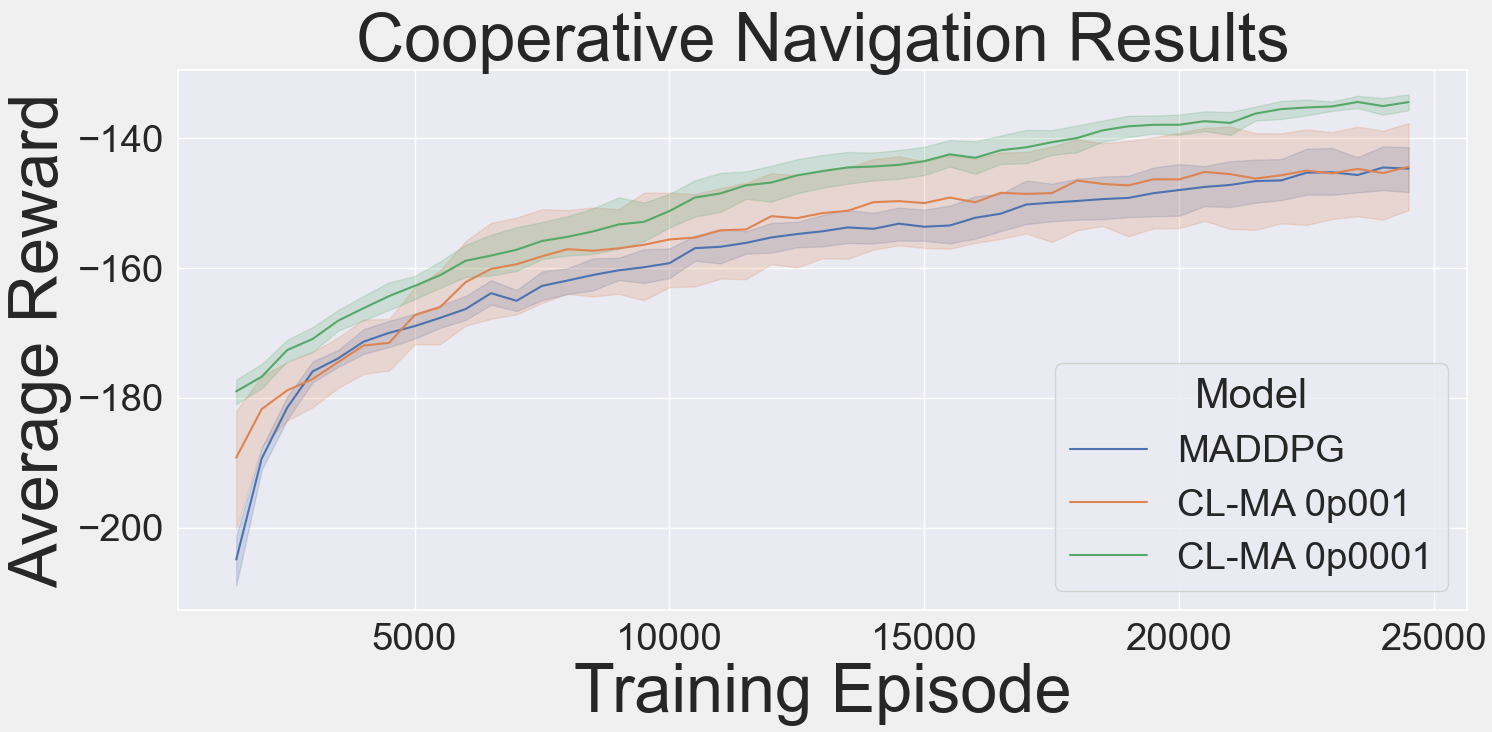}
    \caption{MADDPG and CL-MADDPG (labelled as CL-MA) training results in the Cooperative Communication environment. Green line shows a CL-MA agent with a $\beta$ coefficient of 1e-3. Orange line shows the same scenario with a coefficient of 1e-2. Blue line shows the traditional MADDPG agent. Averages are shown over 5 seeds, with a rolling average window of 5 episodes used to smooth the curve. Error bars represent 99\% confidence interval.}
    \label{fig:CoopNavigation}
    \end{centering}
\end{figure}

\subsection{Competitive Environment}
\begin{figure}[ht]
    \begin{centering}
    \includegraphics[width=0.75\textwidth]{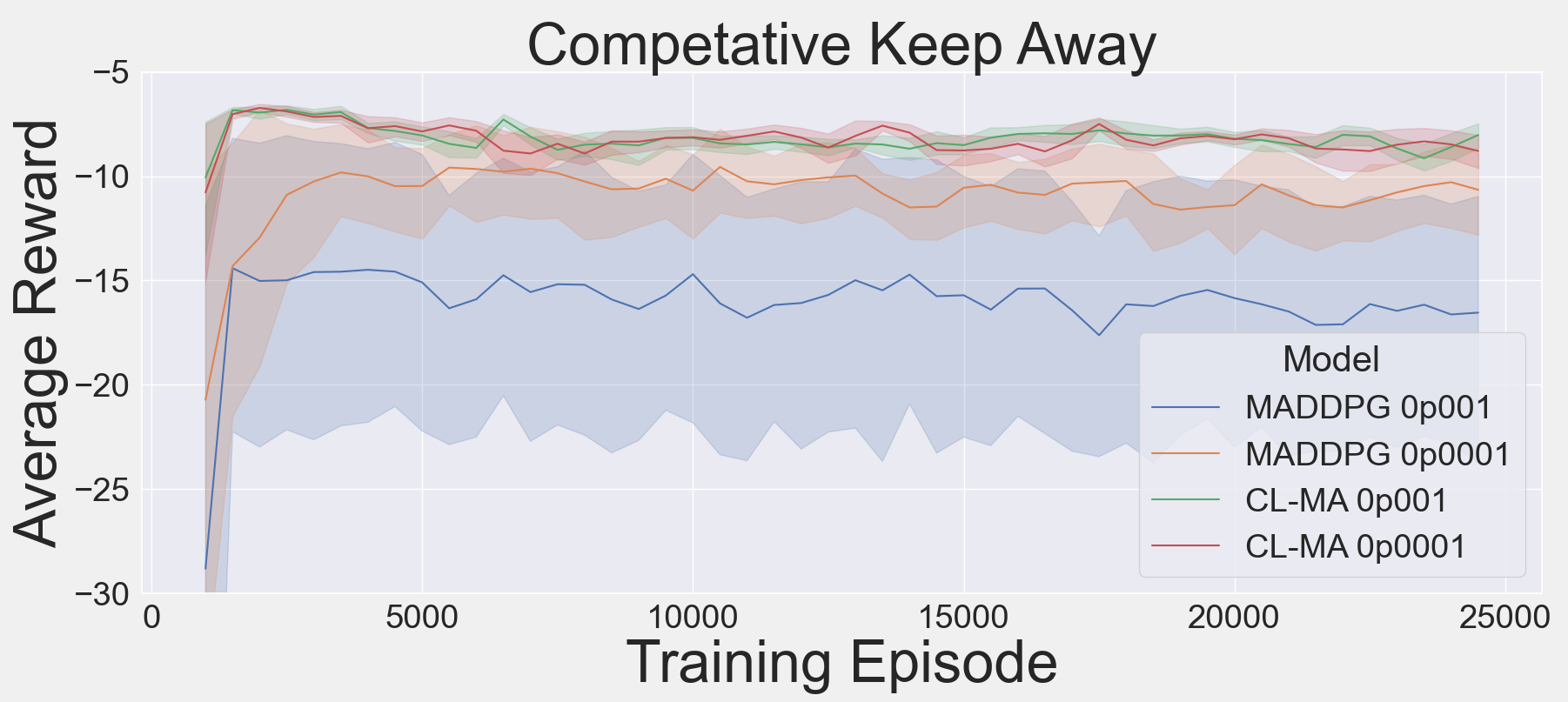}
    \caption{Good MADDPG vs Adversarial CL-MADDPG and vice versa training results in the competitive Push environment. All results report the reward of the `good' agent. Blue represents a good CL-MADDPG agent with $\beta$ coefficient of 1e-2 against a traditional MADDPG agent. Orange represents the same scenario with a 1e-3 coefficient. Red represents a good MADDPG agent against a CL-MADDPG agent with a $\beta$ coefficient of 1e-3. Green represents the same scenario with a coefficient of 1e-2. Averages are shown over 5 seeds, with a rolling average window of 5 episodes used to smooth the curve. Error bars represent 99\% confidence interval.}
    \end{centering}
\label{fig:Push}
\end{figure}

Results from this 1 on 1 competitive environment demonstrate that the capacity-limited version of DDPG when acting as the good agent achieves high reward that remains consistently high throughout training. Meanwhile, when the DDPG agent is matched against an adversary with a capacity-limit on policy information there is a slight decrease in performance in the case of the 1e-3 $\beta$ coefficient agent, and a larger decrease observed when matched against the 1e-2 $\beta$ coefficient. 

\subsection{Mixed Cooperative and Competitive}
\begin{figure}[ht]
    \begin{centering}
    \includegraphics[width=0.75\textwidth]{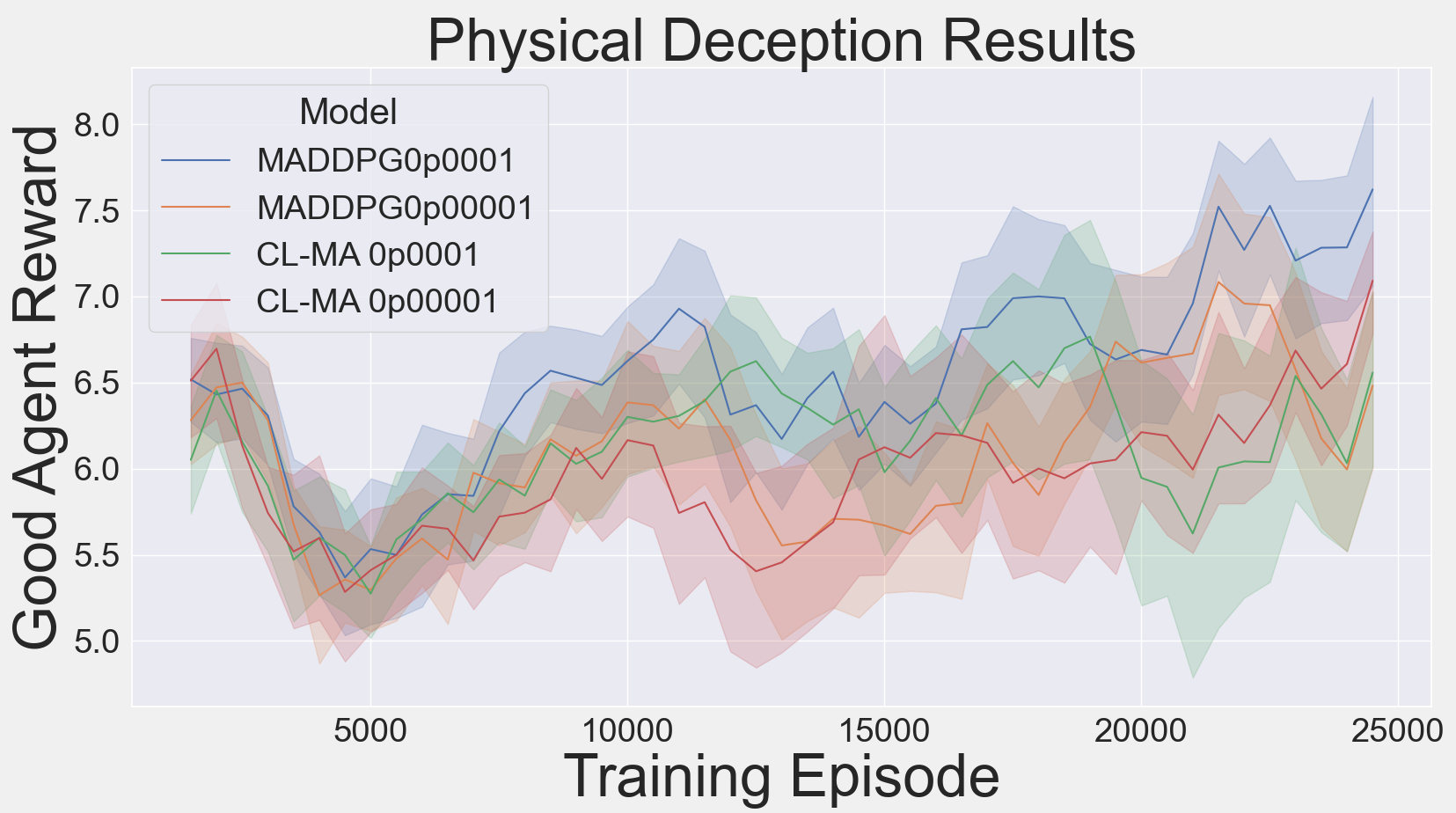}
    \caption{Good MADDPG vs Adversarial CL-MADDPG and vice versa training results in the physical deception environment. All results report the reward of the `good' agent, for MADDPG agents the beta coefficient of the adversary is listed next to the model name in the legend. Colors are represented in the same manner as in Figure 3. Averages are shown over 5 seeds, with a rolling average window of 5 episodes used to smooth the curve. Error bars represent 99\% confidence interval.}
    \end{centering}
\label{fig:Adversary}
\end{figure}

The pattern of the average reward of the good agent in the mixed cooperative and competitive task demonstrates the large amount of nonstationary as it increases and decreases several times throughout training. Here, the capacity-limited version of the DDPG agent with a $\beta$ coefficient of 1e-3 when matched up against a traditional DDPG agent achieves roughly the same performance as the DDPG agent when matched against either the  $\beta$ 1e-3 or $\beta$ 1e-4 capacity-limited agents. The only large deviation of average reward observed in the good agent in this environment occurs when the beta coefficient is large enough to negatively impact performance when $\beta$ is 1e-3. 

This difference in the effect of information capacity limits demonstrates one aspect of this method, in which different environments that have a fairly similar setup can require different amounts of information to represent agents behaviour. Although each of the strictly cooperative or competitive environments showed improved performance through the capacity-limited method, this was not observed in the mixed cooperative and competitive environment. This may be due to the specific dynamics of the physical deception task, or potentially due to the general learning structure of mixed cooperative and competitive tasks.

\section{Conclusions}
In this paper we present a modification of the existing Multi Agent Deep Deterministic Policy Gradient (MADDPG) method by altering the learning objective with a regularization term that limits policy information. The motivation for investigating the impact of this policy information minimization is in improving generalization as well as connections to consolidation of learned behaviour in nonstationary environments. Results from experimentation within these multi-agent cooperative, competitive, and mixed cooperative-competitive tasks indicate that the capacity-limited approach is a good candidate for improving generalization and consolidation of learning in nonstationary environments.

Several areas of future research exist within the understanding of how attempts of generalization impact the consolidation of learned behaviour in nonstationary environments. Although the capacity-limited method showed improvements in learning for most of these tasks, the physical deception task did not show a significant different. It is unclear precisely what dynamics of this learning task makes the capacity-limited approach less successful. Other environments tested included ones with more than 2 agents, and environments that had cooperative or competitive dynamics. This was the only environment tested that had a mix of cooperative and competitive aspects within the same learning environment. Future research in applying capacity-limits to mixed cooperative and competitive tasks to determine if this effect is common across these types of environments.

\bibliographystyle{unsrt}  
\bibliography{references}  

\end{document}